\pdfoutput=1

\documentclass[11pt]{article}

\usepackage[]{acl}

\usepackage{times}
\usepackage{latexsym}

\usepackage{amsmath}
\usepackage{graphicx}
\usepackage{amsfonts}
\usepackage{float}
\usepackage{booktabs}
\usepackage{comment}
\usepackage[normalem]{ulem}

\usepackage{caption, subcaption}

\usepackage{tcolorbox}
\usepackage{color, colortbl}

\definecolor{LRed}{rgb}{1,.8,.8}
\definecolor{MRed}{rgb}{1,.6,.6}
\definecolor{HRed}{rgb}{1,.2,.2}

\definecolor{LGreen}{rgb}{.8,1,.8}
\definecolor{MGreen}{rgb}{.6,1,.6}
\definecolor{HGreen}{rgb}{.2,1,.2}

\usepackage[T1]{fontenc}

\usepackage[utf8]{inputenc}

\usepackage{microtype}

\usepackage{inconsolata}

\usepackage{graphicx}

\title{Human Speech Perception in Noise: Can Large Language Models Paraphrase to Improve It?}

\author{Anupama Chingacham\textsuperscript{1} \,
 Miaoran Zhang\textsuperscript{1} \, Vera Demberg\textsuperscript{1,2} \, \textbf{Dietrich Klakow}\textsuperscript{1} \\
\textsuperscript{1}Saarland University, Saarland Informatic Campus, Germany \\
\textsuperscript{2}Max Planck Institute for Informatics, Germany\\
{\tt  achingacham@lsv.uni-saarland.de}
}

\begin{document}
\maketitle

\begin{abstract}

Large Language Models (LLMs) can generate text by transferring style attributes like formality resulting in formal or informal text.
However, instructing LLMs to generate text that when spoken, is more intelligible in an acoustically difficult environment, is an under-explored topic.
We conduct the first study to \textit{evaluate LLMs on a novel task of generating acoustically intelligible paraphrases} for better human speech perception in noise.
Our experiments in English demonstrated that with standard prompting, LLMs struggle to control the non-textual attribute, \textit{i.e.,} acoustic intelligibility, while efficiently capturing the desired textual attributes like semantic equivalence. 
To remedy this issue, we propose a simple prompting approach, \textit{prompt-and-select}, which generates paraphrases by decoupling the desired textual and non-textual attributes in the text generation pipeline.
Our approach resulted in a $40\%$ relative improvement in human speech perception, by paraphrasing utterances that are highly distorted in a listening condition with babble noise at signal-to-noise ratio (SNR) $-5$ dB. This study reveals the limitation of LLMs in capturing non-textual attributes, and our proposed method showcases the potential of using LLMs for better human speech perception in noise.\footnote{Our code and data are available at \url{https://github.com/uds-lsv/llm_eval_PI-SPiN}.}


\end{abstract}

\section{Introduction}
Paraphrase generation is the task of rephrasing a sentence while retaining its meaning \cite{bhagat2013paraphrase}. Humans perform paraphrasing in spoken conversations, to enable their listeners to perceive spoken messages as intended \cite{bulyko2005error, bohus2008sorry}. Motivated by human speech production strategies, paraphrasing has also been applied to speech synthesis systems, to enhance the quality, naturalness \cite{nakatsu2006learning, boidin2009predicting}, and intelligibility of synthetic speech, especially in challenging acoustic conditions \cite{zhang2013rephrasing}.
Recent explorations on why certain sentences are more intelligible than their paraphrases showed that, the observed intelligibility gain in a noisy listening environment is attributed to the rephrasing, which introduces more acoustic cues that survived the masking effect of the noise \cite{chingacham21_interspeech, chingacham2023data}. 
In other words, the enhanced speech perception with paraphrasing is driven by noise-robust acoustic cues.

The potential of paraphrasing is however, seldom used to build human-like spoken dialogue systems that are adaptive to human listeners' perception errors in noise, presumably due to the limited investigations to generate paraphrases that are acoustically more intelligible in a noise condition.
Prior studies relied on human annotations to identify the ideal paraphrase among a set of candidates \cite{nakatsu2006learning, zhang2013rephrasing, chingacham2023data}, with little discussion on generating intelligible paraphrases. 
This raises the question of \textit{how to generate text that is semantically equivalent to and acoustically more intelligible than the given input sentence, for a noisy environment}. We refer to this task as \textbf{Paraphrase to Improve Speech Perception in Noise} (PI-SPiN).

\begin{figure*}[t]
\centering
  \includegraphics[width=0.90\linewidth]{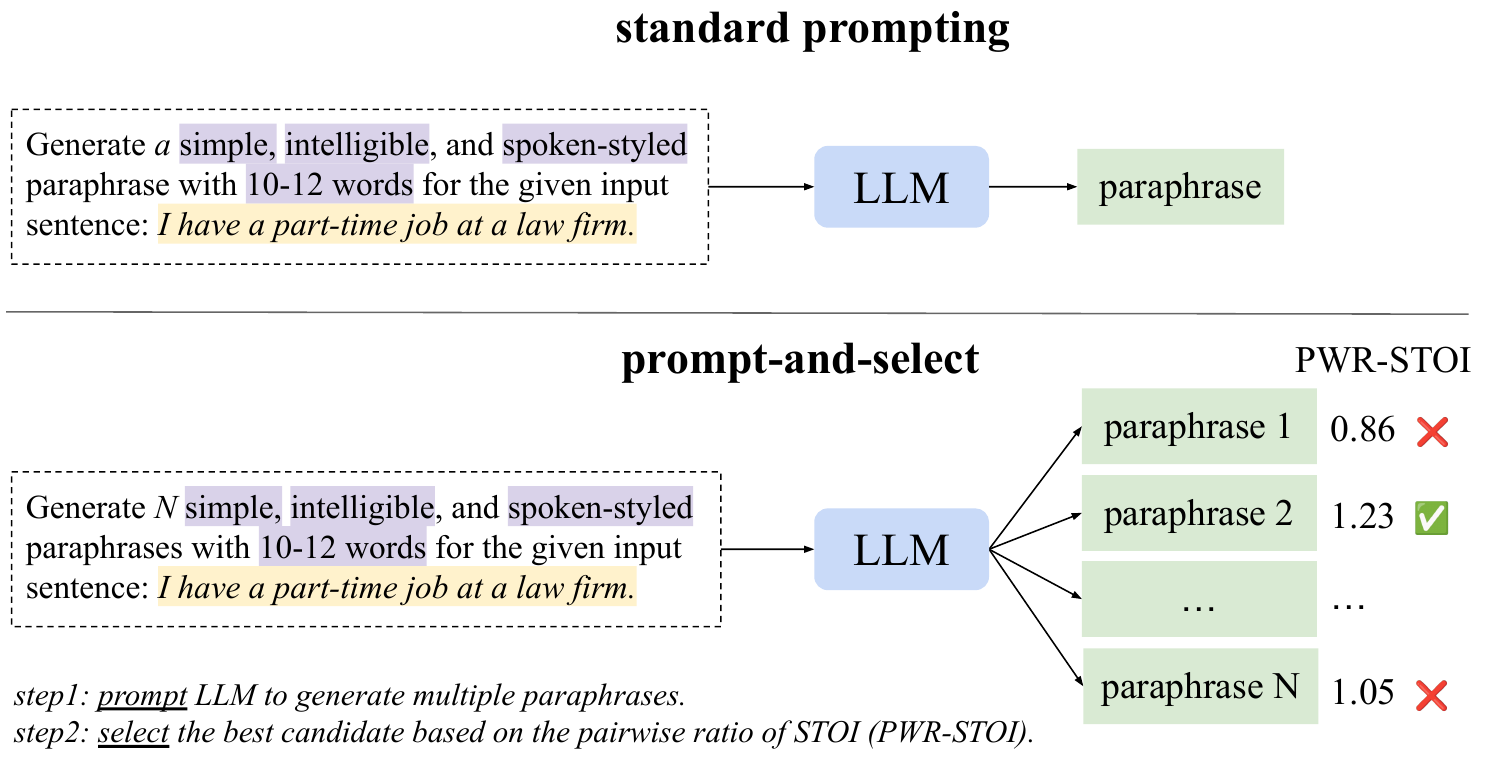}
  \caption{A schematic representation of the \textit{prompt-and-select} and standard prompting approach to generate acoustically intelligible paraphrase in a noisy environment. A speech intelligibility metric, short-time objective intelligibility measure (STOI) is employed to select the paraphrase that is more likely to improve speech perception. \looseness-1
  }
  \label{fig: pas_overview}
  \vspace{-2mm}
\end{figure*}

This task is particularly interesting in the context of generative LLMs, which have shown incredible performance in natural language generation (NLG) tasks such as paraphrase generation and dialogue generation \cite{radford2019language, wei2022finetuned, li2024pretrained}. 
Moreover, recent studies have demonstrated LLMs' capability to control text generation for a wide range of style attributes like sentiment, syntax, formality, and politeness \cite{zhang2023survey, sun-etal-2023-evaluating}. PI-SPiN differs from those controllable text generation problems, as it aims to generate text that satisfies the desired textual attributes (e.g., semantic equivalence), in addition to the non-textual attribute (\textit{i.e.} acoustic intelligibility), which is hard to describe textually.

To explore the potential of LLMs in PI-SPiN, we proposed to \textit{evaluate LLMs' inherent capability to generate acoustically intelligible paraphrases}, without any model fine-tuning.
Through standard prompting methods like zero-shot learning (ZSL), we found that the model was able to capture textual attributes, while consistently struggling to improve acoustic intelligibility. 
We also observed that increasing the description of the desired non-textual attribute in the prompt only confuses the model, and it may even lead to a deterioration in textual attributes that were achievable otherwise. 

To effectively utilize LLMs for generating acoustically intelligible paraphrases, we propose a simple approach called \textit{prompt-and-select}, which guides paraphrase generation by introducing the desired non-textual attribute in a post-processing step (see Figure~\ref{fig: pas_overview}).
It is a two-step process beginning with prompting the LLM to generate multiple candidates and then selecting the best candidate based on acoustic intelligibility, which is hard to capture in textual mode alone.
By conducting a human evaluation with native English listeners, who have no hearing impairments, we verified that the LLM-generated paraphrases via \textit{prompt-and-select} approach are indeed more intelligible than original sentences, in a listening environment with babble noise at SNR~$-5$ dB.\footnote{See definitions of babble noise and SNR in Appendix~\ref{sec: noise}.}

Our main contributions are as follows:

\begin{itemize}
    \item We conduct an elaborate study on the evaluation of LLMs on a novel task called PI-SPiN.
    \item Our results illustrate the weakness of standard textual prompting to control a non-textual attribute -- acoustic intelligibility.
    \item Our proposed approach \textit{prompt-and-select} is an effective solution to generate paraphrases that are more acoustically intelligible.
\end{itemize}

\section{Related Work}


\paragraph{Acoustic Intelligibility.}

Speech perception has been a long-standing research topic in speech science \cite{kalikow1977development, luce1998recognizing, mcardle2008predicting}, which contributed towards a better understanding of human (\textit{mis})hearing. More specifically, several human perception experiments were conducted to investigate the intelligibility of speech tokens such as vowels \cite{doi:10.1121/1.1908983,doi:10.1121/1.1810292}, consonants \cite{weber2003consonant,doi:10.1121/1.3224721}, words in isolation \cite{luce1998recognizing, lexicalneighborhoods, wilson2008comparison}, words in context \cite{kalikow1977development,  uslar2011does, chingacham21_interspeech}, especially in noisy environments.
While several studies showcased the influence of linguistic characteristics such as predictability \citep{kalikow1977development}, syntactic complexity \citep{uslar2011does, carroll2013effects, van2018speech}, and lexical features \citep{luce1998recognizing, mcardle2008predicting}, on the intelligibility of utterances in noise, there is limited explorations in utilizing the linguistic potential to improve acoustic intelligibility in noise. 

We share the motivation to improve speech perception in noise using paraphrases with early studies \cite{cox2004modelling, nakatsu2006learning, zhang2013rephrasing, chingacham2023data}.
\citet{nakatsu2006learning} proposed to train a re-ranker to select the paraphrases that are more likely to sound natural, when synthesized. They generated multiple paraphrases for each sentence mainly by modifying the word order and replacing a few lexical units in the original sentence. 
On the other hand, \citet{zhang2013rephrasing} proposed an objective measure to distinguish the intelligibility difference among paraphrases that are of the same syntactic type, thereby restricting the type of sentential paraphrases.
More recently, \citet{chingacham2023data} investigated the potential of improving intelligibility by considering a larger set of paraphrase types, which are generated using modern paraphrasing models.
However, our work is distinct from theirs as we explore LLMs' inherent ability to generate acoustically intelligible paraphrases.
\raggedbottom

\paragraph{LLM Evaluation.}

Given the rapid growth of LLMs such as ChatGPT and GPT-4~\citep{openai2023gpt4}, there has been a surge of research interest towards a holistic evaluation of their capabilities~\citep{chang2024survey}. Recent studies have attempted to examine their performance across diverse tasks, such as machine translation~\citep{hendy2023good, zhu2023multilingual}, text summarization~\citep{yang2023exploring, pu-demberg-2023-chatgpt}, etc; and also aspects of multilinguality~\citep{lai2023chatgpt, ahuja-etal-2023-mega} and multimodality~\citep{bang2023multitask}. Close to our work, there have been a few studies looking into the controllable generation ability of LLMs. \citet{lai2023multidimensional} explore the potential of ChatGPT as a text-style transfer evaluator. \citet{sun2023evaluating} present a systematic study on ten controllable generation benchmarks. Notably, their control factors are derived from the language perspective (e.g., sentiment and number), whereas our work pioneers the investigation of the potential of LLMs as an acoustically intelligible paraphrase generator.

\section{PI-SPiN Task Description}

Typically, the paraphrase generation task focuses on generating text that is semantically equivalent to the given input text. However, the PI-SPiN task aims at generating text that is semantically equivalent to, as well as, acoustically more intelligible than the original input text, in an adverse listening condition. 

For example, consider the following paraphrase triplet ($s_1, s_2, s_3$) from the Paraphrases-in-Noise (PiN) dataset\footnote{See Appendix~\ref{sec: more_pin} for more samples.} \cite{chingacham2023data}:

\begin{itemize}
    \item[$s_1$:] \textit{``i was raised in a generation we did need all those things.''}
    \item[$s_2$:] \textit{``we did need all those things when i was a child.''} 
    \item[$s_3$:] \textit{``we did need all those things when i was young.''} 
\end{itemize}

$s_1$ is a sentence retrieved from a spoken corpus, while $s_2$ and $s_3$ are outcomes of a paraphrase generation pipeline.
Though all sentences are semantically equivalent to each other, they exhibited a significant difference in acoustic intelligibility under noise. 
More precisely, when these sentences were uttered in a difficult listening condition with babble noise at an SNR of $-5$ dB, humans perceived $s_2$ with fewer errors in perception compared to $s_1$, while $s_3$ was perceived much worse than $s_1$. The better intelligibility of utterances can be attributed to both linguistic features like predictability \cite{kalikow1977development}, syntactic structure \cite{uslar2013development}, as well as acoustic features like the underlying sounds of the utterance \cite{luce1998recognizing}. In the more recent investigations on the intelligibility difference among paraphrases \cite{chingacham2023data}, it was shown that the better intelligibility of $s_2$ in such high noise environments is mainly driven by the noise-robust acoustic cues that are defined by both the constituting sounds as well as the noise signal.
PI-SPiN aims to generate paraphrases (like $s_2$) that are likely to improve human speech perception in such noisy conditions.

Speech intelligibility in noise is better when sentences are simple \cite{carroll2013effects}, shorter \cite{coene2016linguistic}, and linguistically more predictive \cite{valentini2014using}. However, the intelligibility of an utterance in noise is not only driven by its underlying text. The perception is also influenced by the acoustic cues that survived the masking effect of the background noise \cite{cooke2006glimpsing}. Hence, PI-SPiN is a text generation task, that involves both textual attributes like semantic equivalence and a non-textual attribute that captures the noise-robustness of an utterance.

To generate the acoustic realization of a sentence, we used the Tacotron2 text-to-speech (TTS) system, which demonstrated performance on par with that of a professional voice talent \cite{shen2018natural}.
More specifically, we used the Tacotron2 model\footnote{\url{https://huggingface.co/speechbrain/tts-tacotron2-ljspeech}} pre-trained on the LJSpeech dataset by SpeechBrain \cite{speechbrain}. Further, to create the noise-distorted signals, the clean audio signals underwent a noise-mixing procedure using an open-sourced tool, \textit{audio-SNR}.\footnote{\url{https://github.com/Sato-Kunihiko/audio-SNR}}
The babble noise from the NOISEX-92 dataset \cite{varga1993assessment} was mixed with clean audio at SNR$-5$ dB.
To determine whether the generated text satisfies the desired outcome, we primarily relied on automatic metrics, which are discussed in detail in the following section.

\section{Experimental Setup}
\label{sec: experimental_setup}




\paragraph{Model.} For all our experiments, we used \textit{ChatGPT}\footnote{Version: \texttt{gpt-3.5-turbo}}~\citep{NEURIPS2022_b1efde53}, which is one of the most popular LLMs.
It has shown impressive performance on paraphrase generation with textual style attributes, while its ability on acoustically intelligible paraphrasing remains unclear. We adopt default parameters (temperature=$1.0$, top\_p=$1.0$) for the API calls.

\paragraph{Dataset.}
The evaluation dataset consists of $300$ short sentences, which are spoken in a conversational scenario.
The dataset is created by filtering out sentences with $10$ to $12$ words from the top $1000$ lines of the speech corpus, Switchboard \cite{10.5555/1895550.1895693}.

\paragraph{Metrics.}
Human evaluation is the gold standard for most text-generation tasks. However, human evaluation is expensive and time-consuming, which limits the scale of evaluation. Thus, we perform an automatic evaluation of the whole evaluation dataset and a human evaluation of a subset of the dataset.
For automatic evaluation, we employed a range of metrics, which determine the semantic equivalence between the input and output texts, as well as, the linguistic and acoustic features that contribute to the acoustic intelligibility in noise.


\paragraph{\textnormal{\textit{1. Semantic equivalence.}}} 
Semantic Textual Similarity (STS) measures how similar two texts are in terms of their meaning.
In the past, several STS scores were proposed \cite{bar2012ukp, han2013umbc_ebiquity}.
More recently, \citet{zhang2020bertscore} proposed BERTScore, which has shown 
encouraging results in correctly identifying the semantic equivalence/distance between two texts.
For all our evaluations, the STS score is the BERTScore-f1 calculated using the distilled BERT model \cite{Sanh2019DistilBERTAD}. The higher the STS value, the better the semantic equivalence between two texts.

\paragraph{\textnormal{\textit{2. Lexical deviation.}}}
Lexical deviation (LD) shows to what extent two texts are similar or different in terms of their surface form. The difference in wording between the two texts is particularly interesting for paraphrase generation.  \citet{bandel2022quality} showed that the deviation in the linguistic forms of paraphrases is one of the critical factors that decides its quality -- high-quality paraphrases exhibit high LD, as well as, high STS as they differ lexically, yet maintain the semantics. As defined in \citet{liu2022towards}, we used the overlap in lexical tokens of the uncased lemmatized form of two texts to capture the lexical deviation between the input sentence and the model-generated paraphrase. The higher the LD score, the more difference in paraphrased wording. 

\paragraph{\textnormal{\textit{3. Utterance length.}}}
It is a textual attribute that influences acoustic intelligibility, as it was observed that shorter sentences introduce fewer misperceptions in noise \cite{chingacham2023data}. 
Though paraphrases of shorter lengths are more likely to be perceived correctly, shorter paraphrases may risk missing some semantics of the original text. Hence, it is critical to evaluate utterance length along with semantic equivalence.
To measure utterance length in terms of phonemes (\textit{i.e.} PhLen), we used a grapheme-to-phoneme model\footnote{\url{https://pypi.org/project/g2p-en/}} to generate the phonemic transcript of a sentence. 
Further, to compare the length within each input-output pair, the \textit{pairwise ratio of PhLen} is calculated by dividing the length of the model output by that of its input sentence (denoted as \textit{PWR}-PhLen). 
Thus, when the model-generated text is similar to the input text, \textit{PWR}-PhLen value is close to $1.0$, while a value much less than $1.0$ reflects that the model-generated text is considerably shorter than the original text. \looseness-1

\begin{table*}[!htb]
    \centering
    \scalebox{1.0}{
    \begin{tabular}{cl}
        \toprule
         Prompt-ID & Prompt \\ 
         \midrule
         \midrule
         $p_{zsl-low}$ &  Generate an \textbf{intelligible paraphrase} for the following input sentence: \texttt{\{input text\}} \\
         \midrule
         $p_{zsl-med}$ 
         &  Generate a \textbf{simple}, \textbf{intelligible}, and \textbf{spoken-styled paraphrase} with \textbf{10-12 words} \\
         & for the following input sentence: \texttt{\{input text\}} \\
         \midrule
         $p_{zsl-high}$  &  \textbf{For a noisy listening environment with babble noise at SNR -5}, generate a \textbf{simple}, \\
         & \textbf{intelligible}, and \textbf{spoken-styled paraphrase} with \textbf{10-12 words}, for the following input \\
         &sentence: \texttt{\{input text\}}\\
    \bottomrule
    \end{tabular}}
    \caption{Three prompts used in standard prompting, with an increasing level of detail in the task objective. Bold-faced words are task-specific keywords in the prompt statement. 
    }
    \label{tab:prompt_zsl}
\end{table*}

\paragraph{\textnormal{\textit{4. Linguistic predictability.}}}
Several studies in the past have shown that when lexical tokens are more predictable from the context, word misperceptions are less likely to occur \cite{kalikow1977development, uslar2013development, valentini2014using, schoof2015high, 10.3389/fpsyg.2021.714485}. 
Thus, we considered the perplexity (PPL) score determined by a pre-trained language model, GPT-2\footnote{Version: \texttt{distilgpt2}} \cite{radford2019language} to estimate the linguistic predictability of a sentence. 
To compare the linguistic predictability among input and output texts, the \textit{pairwise ratio of the perplexity} is calculated by dividing the PPL of model-generated text by the input sentence PPL (denoted as \textit{PWR}-PPL). Higher PPL scores indicate lesser linguistic predictability. Thus, a \textit{PWR}-PPL value less than $1.0$ indicates that the model-generated text is more predictable than the input text. 

\begin{table*}[!htb]
\centering
\setlength{\tabcolsep}{10pt}
\begin{tabular}{l|lllll}
\toprule
Prompt-ID &  STS$\uparrow$ & LD$\uparrow$ & \textit{PWR-}PhLen$\downarrow$ & \textit{PWR-}PPL$\downarrow$ & \cellcolor{LGreen}\textit{PWR-}STOI $\uparrow$ \\ \midrule
\midrule
$p_{zsl-low}$ & 0.852 & 0.699 & 1.343* & 1.086 & \cellcolor{LGreen}0.992 \\
$p_{zsl-med}$ &  0.860 & 0.668 & 1.119* & 1.042 & \cellcolor{LGreen}0.991 \\
$p_{zsl-high}$ &  0.837 & 0.729 & 1.250* & 1.236* & \cellcolor{LGreen}1.005 \\ 

\bottomrule
\end{tabular}
\caption{An automatic evaluation of paraphrases generated by different prompts. Pairwise ratios (\textit{PWR}) significantly different from $1.0$ ($p < 0.05$) are marked with an asterisk (\textbf{*}). They indicate the significant difference between the model-generated output and the input text.
}
\label{tab: evaluation on zsl}
\end{table*}


\paragraph{\textnormal{\textit{5. Acoustic Intelligibility.}}}
The acoustic intelligibility of an utterance in a noisy environment is primarily driven by the acoustic cues that survived the energetic masking of the noise -- utterances with better noise-robust acoustic cues are better perceived in noise \cite{cooke2006glimpsing, tang16c_interspeech}.
We use the Speech Intelligibility (SI) metric, STOI \cite{taal2010short}, to capture the acoustic intelligibility of an utterance. 
STOI is a non-textual attribute, as it measures the mean correlation of short-time envelopes between the clean and noisy audio signals of an utterance. The higher the STOI score, the higher the noise-robustness of an utterance.
Similar to other pairwise ratios, the \textit{pairwise ratio of STOI} (\textit{PWR}-STOI) is calculated by dividing the STOI of model-generated text by the input text STOI. Thus, PI-SPiN aims at generating paraphrases with \textit{PWR}-STOI values above $1.0$ indicating that the model output is acoustically more intelligible than the input sentences. 

All pairwise ratios range between $0.0$ and $+\infty$, while STS and LD range between $0.0$ and $1.0$. For the evaluation, we report each of these metrics, averaging across the evaluation dataset.

\section{Evaluating LLMs for PI-SPiN}
\label{sec: evaluating_llm}

In our experiments, an LLM is prompted to generate a paraphrase for each input sentence in the evaluation set with a prompt template: \texttt{\{prompt prefix\} +  \{input text\}}. 
In the following section, we described two prompting methods that we employed and evaluated for the task.

\subsection{Standard Prompting} 
In this setting, the model is prompted to generate an intelligible paraphrase given an input sentence in a zero-shot manner.   
As shown in Table~\ref{tab:prompt_zsl}, we investigate three types of prompts, which describe the desired attributes with different granularity: low ($p_{zsl-low}$), medium ($p_{zsl-med}$), and high ($p_{zsl-high}$). With the increasing number of task-specific tokens in the prompt, the task description is more detailed. Prompts are designed by including keywords like `\textit{paraphrase}' and `\textit{intelligible}' that represent the desired outcome.
Additionally, a few more tokens like `\textit{10-12 words}' and `\textit{spoken-styled}' were used in the prompt to ensure that the generated paraphrase adheres to the length and style of input sentences. We hypothesize that with additional task-oriented tokens in the prompt, the model will steer the paraphrase generation by optimizing the intelligibility.
\begin{table*}[!htb]
    \centering
    \scalebox{0.95}{
    \begin{tabular}{c|cl}
    \toprule
    Prompt-ID & \multicolumn{2}{c}{Sentences} \\
    \midrule \midrule
    ---& \texttt{\{input text\}} & \textit{I also have a part-time job at a law firm.}     \\
    $p_{zsl-low}$ & \texttt{\{output\}} & \textit{Additionally, I am employed part-time at a law firm.} \\
    $p_{zsl-med}$ & \texttt{\{output\}} & \textit{I work part time at a law firm as well.} \\
    $p_{zsl-high}$ & \texttt{\{output\}} &  \textit{\textbf{In a busy environment with loud chatter}, I work part-time at a law office.} \\
    \midrule
    ---& \texttt{\{input text\}} & \textit{Is your father's cafe, one that you could cater out of?}   \\
    $p_{zsl-low}$ & \texttt{\{output\}} & \textit{Can you use your father's cafe as a location for catering purposes?} \\
    $p_{zsl-med}$ & \texttt{\{output\}} & \textit{Could you use your father's cafe as a catering venue?} \\
    $p_{zsl-high}$ & \texttt{\{output\}} &  \textit{Can your dad's café be used for catering \textbf{in a noisy setting}?}    \\
    \midrule
    ---& \texttt{\{input text\}} & \textit{I believe that the cure for polio is already founded. } \\
    $p_{zsl-low}$ & \texttt{\{output\}} & \textit{I hold the belief that the remedy for polio has already been discovered.}  \\
    $p_{zsl-med}$ & \texttt{\{output\}} & \textit{I think the polio cure has already been discovered.} \\
    $p_{zsl-high}$ & \texttt{\{output\}} &  \textit{I think polio's cure has already been discovered \textbf{in a noisy setting.}}\\ 
    \midrule
    ---& \texttt{\{input text\}} & \textit{I think you can do that for a little while.}   \\
    $p_{zsl-low}$ & \texttt{\{output\}} & \textit{I believe you have the capability to do that temporarily.} \\
    $p_{zsl-med}$ & \texttt{\{output\}} & \textit{I believe you can manage that temporarily.} \\
    $p_{zsl-high}$ & \texttt{\{output\}} &   \textit{I believe you can manage it temporarily \textbf{amidst the loud chatter.}} \\ \bottomrule
    \end{tabular}}
    \caption{A qualitative analysis of model-generated text, \texttt{\{output\}}, for a given \texttt{\{input text\}} under three standard prompts: $p_{zsl-low}, p_{zsl-med}, p_{zsl-high}$. The prompt
    $p_{zsl-high}$ generates several tokens that are irrelevant (bold-faced words) to the task objective.
    }
    \label{tab:zsl-high_qualitatve_analysis}
\end{table*}

\paragraph{\textbf{Results and Analysis}} \mbox{Table \ref{tab: evaluation on zsl}} summarizes the results of all three prompts that we used in standard prompting.
We observed that ChatGPT can generate high-quality paraphrases as indicated by high scores for semantic equivalence and lexical deviation (\textit{i.e.} STS and LD) See Appendix~\ref{sec: absolute_score} for the absolute scores of different metrics.
More importantly, we found that the length of paraphrases generated by the prompt $p_{zsl-med}$ (PhLen = $42.08$) is considerably shorter than those generated with the prompt $p_{zsl-low}$ (PhLen = $50.67$), indicating the effectiveness of additional keywords in $p_{zsl-med}$ to control a textual attribute -- length. 
However, the non-textual attribute, acoustic intelligibility (\textit{i.e.} STOI) of model-generated paraphrases is not significantly different from their corresponding input sentences as reflected by the \textit{PWR}-STOI scores being not significantly different from $1.0$. 
Furthermore, paraphrases generated with a detailed task description in $p_{zsl-high}$, also resulted in a similar observation -- \textbf{LLM struggles to improve the non-textual attribute while controlling textual attributes appropriately}.



Compared to $p_{zsl-low}$ and $p_{zsl-med}$, $p_{zsl-high}$ resulted in worse performance, indicated by considerably longer output texts despite prompting to control length (\textit{PWR}-PhLen =  $1.250$) and output texts that are linguistically less predictive (\textit{PWR}-PPL =  $1.236$). It is also reflected in a higher lexical deviation (LD = $0.723$) at the expense of lower textual similarity between input and output (STS =  $0.837$). To have a deep understanding of its behavior, we conducted a qualitative analysis as shown in Table~\ref{tab:zsl-high_qualitatve_analysis}. We noticed that the \textbf{additional context of the non-textual attribute confused the model in understanding the task objective and resulted in model hallucination}. In sum, using standard prompting may not effectively elicit the model's ability to generate paraphrases with the intended non-textual attribute, which is beyond the model's comprehension.\footnote{In Appendix~\ref{sec:icl}, we also conducted a preliminary study on in-context learning, suggesting that demonstrations are not helpful in capturing the non-textual attribute.}

\begin{figure*}[!htb]
    \centering
    \begin{subfigure}[b]{0.39\linewidth}
    \centering
    \includegraphics[width=\linewidth]{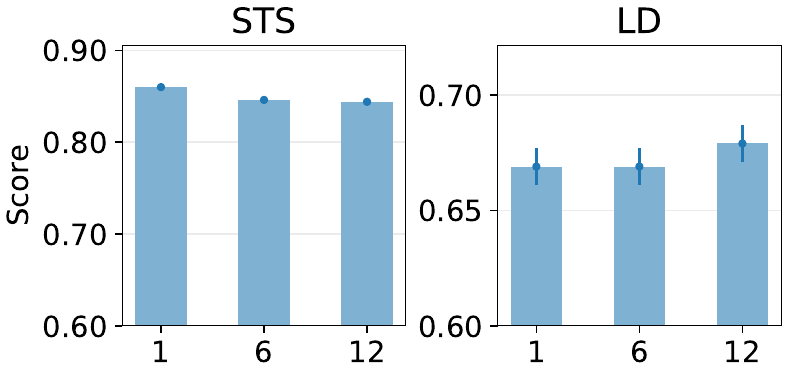}
    \caption{}
    \label{fig:prompt_select_result_a}
    \end{subfigure}
    \hfill
    \begin{subfigure}[b]{0.585\linewidth}
    \centering
    \includegraphics[width=\linewidth]{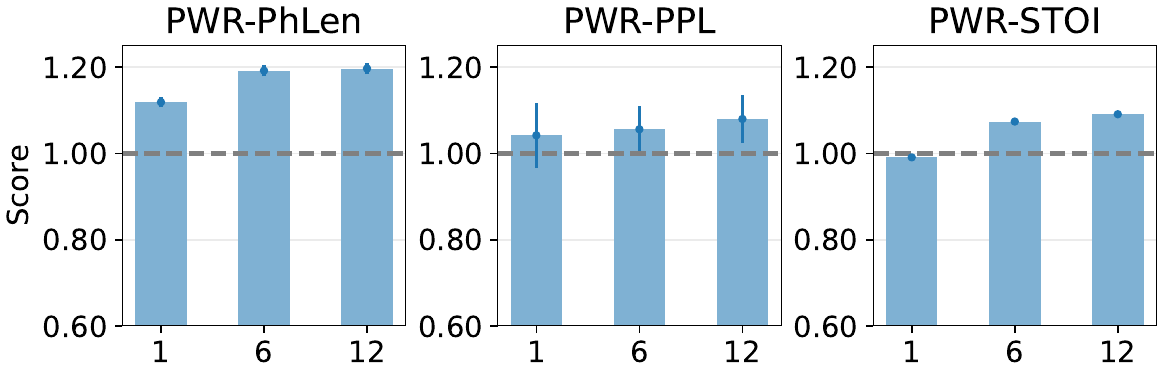}
    \caption{}
    \label{fig:prompt_select_result_b}
    \end{subfigure}
    \vspace{-3mm}
    \caption{An automatic evaluation of the standard prompting ($n=1$) and the proposed prompt-and-select ($n>1$) approach. The \textit{X}-axis is the number of candidates generated ($n$) and the \textit{Y}-axis is the mean scores (with error bars at $95\%$ confidence interval). 
    The reference line in Fig. (b) marks when the input text feature is the same as the output text feature. Increasing $n$ improves the pairwise ratio of acoustic intelligibility (\textit{PWR-}STOI), but it comes with a trade-off on semantic equivalence (STS).
    }
    \label{fig:prompt_select_result}
\end{figure*}


\subsection{PAS: Prompt-and-Select}
Prior studies on dialogue generation \cite{boidin2009predicting, nakatsu2006learning, weston-etal-2018-retrieve} have demonstrated the utility of a simple yet effective pipeline of controlling text generation in two steps: first generating a candidate set of dialogues, and then selecting the best candidate based on the task requirement. Similarly, we proposed to decompose the current task into a two-step process: (1) \textbf{prompt} the LLM to generate multiple output texts that are semantically equivalent to the input text and (2) \textbf{select} the best candidate based on the acoustic intelligibility. 


Our approach is similar to the \textit{prompt-and-rerank} method proposed in \cite{suzgun-etal-2022-prompt}. However, our approach deviates from theirs mainly in two ways: (1) instead of using beam search at the decoding phase, we propose to utilize the potential of an LLM to generate multiple ($n$) candidates that exhibit the desired textual attributes and (2) the best candidate selection is based on a metric (\textit{i.e.} \textit{PWR}-STOI) that represents a non-textual attribute, which is not considered in prior studies. 

For the first step of paraphrase generation, we perform zero-shot prompting with an appropriate task description, $p_{zsl-med}$. Thus, $p_{zsl-med}$ is the prompt that generates exactly one candidate and involves no selection; it is also referred to as $p_{pas(n=1)}$. However, to generate multiple paraphrases (eg: $n=6$), the prompt statement can be simply modified to include the $n$ value, as shown below
\begin{itemize}
    \item \textit{Generate \textbf{6} simple, intelligible, and spoken-styled paraphrases with 10-12 words for the given input sentence: \{\texttt{input text}\}}
\end{itemize}



Following the creation of the candidate set, STOI scores are calculated for all model-generated text as well as the input text, by first synthesizing the clean utterances and then mixing babble noise at SNR~$-5$ dB. Finally, the candidate with the highest \textit{PWR}-STOI is selected as the model output.


\paragraph{Results and Analysis}

We begin our analysis by comparing the results of standard prompting ($n=1$) with the PAS approach, involving $6$ candidates ($n=6$). 
As shown in Figure~\ref{fig:prompt_select_result_a}, PAS showcased a high quality of paraphrase generation as indicated by high STS and high LD, similar to the standard prompting setup.
Similarly, Figure~\ref{fig:prompt_select_result_b} illustrates that other textual attributes like linguistic predictability (\textit{PWR}-PPL = $1.056$) and utterance length (\textit{PWR}-PhLen =  $1.192$) of the PAS approach resulted in similar outcomes of the standard prompting method \textbf{--} output texts are a bit longer than input texts, while their linguistic predictability scores are similar (see Appendix~\ref{sec: absolute_score} for the absolute scores of different metrics with varying numbers of candidates). 
Importantly, compared to the standard prompting, the prompt-and-select approach yielded a noticeably high \textit{PWR}-STOI ($\mu=1.074$, $p<0.05$), which is significantly above $1.0$. This indicates that the model-generated text is considerably more intelligible than their corresponding input sentences in the given noise condition.
We can see more clearly from \mbox{Figure \ref{fig:prompt_select_result_b}} that PAS ($n=6$) leads to a relative improvement of $8.4$\% in \textit{PWR}-STOI compared to the standard prompting ($n=1$). 
Our findings suggest that \textbf{introducing the desired non-textual attribute in a post-processing step is a potential framework to generate desired text with multi-modal behavior}.



\begin{table*}[!htb]
\centering
\begin{tabular}{l|llllll}
\toprule
Subset & STS$\uparrow$ & LD$\uparrow$ & \textit{PWR-}PhLen $\downarrow$ & \textit{PWR-}PPL $\downarrow$ & \textit{PWR-}STOI $\uparrow$ & \cellcolor{LGreen}\textit{PWR-}Sent-Int $\uparrow$\\ \midrule \midrule
$\textrm{top}_{30}$ & 0.831 & 0.737 & 1.189* & 1.428 & 1.22* &  \cellcolor{LGreen} \textbf{1.70*} \\
$\textrm{random}_{30}$ & 0.848 & 0.683 & 1.157* & 1.314 & 1.07* &  \cellcolor{LGreen}{ }1.06 \\ 
\bottomrule
\end{tabular}
\caption{The automatic and human evaluation of text generated with $p_{pas(n=6)}$.
Evaluation on two subsets: top $30$ pairs with highest \textit{PWR}-STOI ($\textrm{top}_{30}$) and randomly selected $30$ pairs ($\textrm{random}_{30}$). \textit{PWR-}Sent-Int captures the pairwise ratio of human speech perception in noise. \textbf{*} marks values significantly above $1.0$ ($p<0.05$).}
\label{tab: overall_evaluation_selected30 on prompts}
\end{table*}

This raises a follow-up question of whether generating more candidates in the first step further improves the overall \textit{PWR}-STOI of generated paraphrases. 
To this end, we modify the number of candidates ($n$) in the prompt statement to double the candidate pool size.
We found that by increasing the candidate set, there is an improvement in acoustic intelligibility.
However, when $n$ is increased from $6$ to $12$, there was only a limited improvement of $1.6$\% in \textit{PWR}-STOI. On the other hand, we observed that textual attributes like linguistic predictability and lexical deviation are not significantly different under varying $n$ values.

Interestingly, the pair-wise ratio of sentence length slightly increased, with more choices in the candidate selection; however, the overall \textit{PWR}-PhLen in this approach is still below the standard prompting setup with no tokens to control length ($p_{zsl-low}$). Increasing $n$ from $6$ to $12$ slightly reduced the overall semantic equivalence between the model input and output paraphrase. This indicates that the choice of $n$ introduces a trade-off between the improvement in acoustic intelligibility (\textit{PWR}-STOI) and the overall semantic equivalence (STS) and one has to choose $n$ considering this trade-off between the gain in non-textual attribute and the need for semantic equivalence. 

\subsection{\textbf{Human Evaluation}}
\label{sec: human_eval}
In addition to the evaluation with automatic metrics, we also conducted a human evaluation to verify whether the model output in the PAS setup (using $p_{pas(n=6)}$) is indeed more intelligible than their corresponding input sentences. For the human perception experiment, we created two subsets of the evaluation dataset of $300$ pairs: $\textrm{random}_{30}$ and $\textrm{top}_{30}$. $\textrm{random}_{30}$ is a set of $30$ pairs randomly selected from the evaluation dataset, while $\textrm{top}_{30}$ is the top $30$ input-output pairs that exhibited the larger improvements in STOI scores.

We followed the experiment design of \citet{chingacham2023data} to capture the human speech perception of an utterance in a (noisy) listening setup. After synthesizing the noisy utterances of each sentence using a TTS \cite{shen2018natural} and a noise-mixing tool (audio-SNR), participants were asked to listen and transcribe each sentence. Every utterance in the dataset was listened to by six different participants. For each listening instance, the edit distance between the phonemic transcriptions of the actual and transcribed text is measured to determine the rate of correct recognition. Furthermore, the sentence-level intelligibility (Sent-Int) of each utterance is calculated by averaging the correct recognition rates exhibited by the six listeners.


The perception experiment was conducted with $24$ native English listeners with no hearing impairments ($14$ females and $10$ males; average age = $30.71$).
After data collection, we calculated the pairwise ratio of sentence-level intelligibility (\textit{PWR}-Sent-Int) by dividing the Sent-Int scores of the output paraphrase by their corresponding input sentence. A mean score of \textit{PWR}-Sent-Int significantly above $1.0$ indicates that the model-generated paraphrase is significantly more intelligible than the input sentence, in a given listening condition.\raggedbottom

\paragraph{Results and Analysis}
As illustrated in Table~\ref{tab: overall_evaluation_selected30 on prompts}, $\textrm{top}_{30}$ items signify that the model-output paraphrases have considerably improved the human perception in a noisy listening condition. We observed that the overall human speech perception of model-output paraphrases (Sent-Int = $0.66$) was considerably higher than the input sentences (Sent-Int = $0.47$), introducing a\textbf{ $40$\% relative gain in the overall intelligibility}. This is also reflected in the \textit{PWR}-Sent-Int score that is significantly above $1.0$.

We observed the \textit{PWR}-Sent-Int of $\textrm{random}_{30}$ is not significantly above $1.0$, even though the \textit{PWR}-STOI is significantly above $1.0$. 
With further analysis of two subsets, we found that the mean STOI of input sentences in $\textrm{top}_{30}$ ($\mu =  0.507$) is significantly less than $\textrm{random}_{30}$ ($\mu =  0.561$).
This means that $\textrm{random}_{30}$ consists of sentences that are better intelligible in noise. Also, we observed a strong negative correlation ($r = -0.442, p <0.001$) between the STOI of input sentences and the gain in intelligibility (\textit{PWR}-Sent-Int), which highlighted the limited benefits of paraphrasing input sentences in $\textrm{random}_{30}$. However, $\textrm{top}_{30}$ consists of all input sentences, which are more likely to benefit from paraphrasing in noisy listening conditions and they reflected the same in the human evaluation. We conclude with the observation PAS is a simple yet effective solution to alleviate the struggles of LLM to generate text with textual and non-textual attributes, without model fine-tuning.

\section{Conclusion}
In this work, we evaluate LLMs on acoustically intelligible paraphrase generation for better human speech perception in noise. Our results demonstrate the limitations of LLMs in controlling text generation with a non-textual attribute -- acoustic intelligibility. 
To alleviate the struggles of LLMs in generating text that satisfies both textual and non-textual attributes, we proposed a simple yet effective approach called \textit{prompt-and-select}.
With human evaluation, we found that when the original utterances are highly prone to misperceptions in noise, prompt-and-select can introduce $40\%$ of relative improvement in human perception. We hope the findings of this work inspire further explorations to control LLMs' text generation with different real-world context cues, thereby building more human-like agents.
For future work, we could consider two approaches to improve LLMs on this task: 1) fine-tuning LLMs with a large parallel dataset consisting of sentences and their corresponding intelligible paraphrases, and 2) incorporating the acoustic representation of the utterances to control the paraphrase generation.


\section*{Limitations}
The proposed ``prompt-and-select'' approach relies on the efficacy of STOI scores to identify the best candidate which is more likely to be perceived correctly in noise. In other words, this approach requires a metric that accurately estimates the desired non-textual attribute. This could be a limitation for problems that require human annotations for candidate selection.
Additionally, the current approach introduces an overhead in computation and inference time, due to multiple generations and STOI calculation that involves speech synthesis and noise-mixing procedure. 
Further investigations are required to study the trade-off between the benefits of paraphrasing and the cost of additional resources.
Moreover, our study only evaluated ChatGPT, one of the representative LLMs, due to budget and resource constraints. We believe that a holistic evaluation covering more open-source models, such as Mistral~\citep{jiang2023mistral7b} and Llama 3~\cite{meta2024introducing}, will be beneficial to deepen our understanding of LLM capabilities.


\section*{Ethics statement}
In this work, generative LLMs are evaluated for a new task without model fine-tuning. It is an impactful step to democratize LLMs for research facilities with limited data and computing resources.
We conducted a human evaluation on Prolific, ensuring that all participants were paid (9 GBP) for their service, considering the recommended minimum wage per hour in the UK, in 2023.
Also, we ensured to provide an inclusive environment for our participants in the perception experiment, providing non-binary options to mark their gender identity.

\section*{Acknowledgements}
We would like to thank anonymous reviewers for their valuable feedback. This work was funded by the Deutsche Forschungsgemeinschaft (DFG, German Research Foundation) – project-id 232722074 – SFB 1102.

%
\bibliography{custom,sfba4,anthology}

\clearpage
\appendix

\section{Definitions}
\label{sec: noise}

\paragraph{Babble Noise.}

It is one of the most commonly occurring noise types in the real world~\cite{miller1947masking}.
Typically, it is the noise that exists in a cafeteria or other crowded environments, wherein individuals engage in conversations in the backdrop of other conversations. The simultaneous speech produced by several individuals in the background masks the target speech and could hinder listening.
The babble noise in the NOISEX-92 database that we use in this work is a recording of $100$ people speaking in a canteen \cite{varga1993assessment, deshpande2009speaker}.

\paragraph{Signal-to-Noise Ratio.}

To measure the noise level, a commonly used metric is the signal-to-noise ratio (SNR)~\cite{taguchi1986introduction}.
SNR represents the ratio of the power of a clean (undistorted) signal and a noise signal, which are combined to form the distorted signal. 
Simply put, it is a fraction of powers as defined in Equation (\ref{eq: snr}). 
It is commonly measured on a logarithmic scale and referred to in units of decibels (dB), as defined in Equation (\ref{eq: snr_db}). 
The power of a signal is the sum of the absolute squares of signal magnitudes averaged across the time domain. 

\begin{equation}
    \label{eq: snr}
    SNR = \frac{P_{signal}}{P_{noise}}
\end{equation}

\begin{align}
    \label{eq: snr_db}
    SNR_{dB} &= 10 \log_{10} (SNR) \nonumber \\
    &= 10 \log_{10} ( \frac{P_{signal}}{P_{noise}} )
\end{align}


When a clean speech signal is mixed with a noise signal with equal power, the SNR of the resultant distorted speech is $0$ dB.
Similarly, when the power of the clean signal is higher than that of the noise, the SNR of the resultant signal is positive ($> 0$ dB). 
Higher SNR scores indicate better audibility. On the other hand, when the noise power is more in the processed signal, the SNR value is negative ($< 0$ dB).

\begin{table}[]
    \centering
    \begin{tabular}{c|cccc}
    \toprule
Prompt-ID & PhLen & PPL & STOI \\ 
\midrule
\midrule
$p_{zsl-low}$ &  50.67 & 159.95 & 0.570  \\
$p_{zsl-med}$ & 42.08 & 165.56 & 0.569 \\
$p_{zsl-high}$ & 46.68 & 193.85 & 0.577  \\ 
$p_{pas_{(n=6)}}$ & 44.67 & 182.77 & 0.617  \\
$p_{pas_{(n=12)}}$ & 44.88 & 184.52 & 0.627  \\ 
$p_{icl}$ & 47.27 & 146.71 & 0.573 \\
\midrule
\texttt{\{input text\}}  & 38.02 & 236.65 & 0.577 \\
\bottomrule
    \end{tabular}
    \caption{Absolute scores for utterance length (PhLen), linguistic predictability (PPL), and acoustic intelligibility (STOI) of \texttt{\{input text\}} and generated outputs by different prompts.}
    
    \label{tab: evaluation_metrics_absolute_scores}
\end{table}

\section{ More Samples from the PiN Dataset}
\label{sec: more_pin}

In Table~\ref{tab: more_examples_PiN_dataset}, we provide more paraphrase triplets from the PiN dataset.




\section{Absolute Scores}
\label{sec: absolute_score}

We provide absolute scores for different evaluation metrics in Table~\ref{tab: evaluation_metrics_absolute_scores} in addition to their pairwise ratios.

\section{In-context Learning} 
\label{sec:icl}

Prior research has shown that LLMs can efficiently learn to control text generation with demonstrations and perform better than just providing a task description \cite{NEURIPS2020_1457c0d6}. 
Thus for the in-context learning (ICL) setup, the input prompt is modified to include a set of exemplars that represent the desired model behavior. 
In other words, to instruct the model to generate acoustically intelligible paraphrases in an ICL setting requires a set of sentences and their corresponding paraphrases that are acoustically more intelligible in a noise condition. 

To provide the best in-context demonstrations, we created another set of $300$ short sentences from the Switchboard corpus excluding those in the evaluation set.
Then, their corresponding paraphrases were generated by prompting ChatGPT with $p_{zsl-med}$. Following speech synthesis and noise mixing with babble noise at SNR~$-5$ dB, we identified the top $5$ pairs that exhibited a larger pairwise difference in STOI scores. Further, the sentences within each pair were rearranged in such a way that the second sentence is always better intelligible than its paired paraphrase.
Further, the sentences within each demonstration pair were concatenated with a token (eg: `=>') and embedded with $p_{zsl-low}$ for in-context learning. 
Table~\ref{tab: prompt icl} represents the exact prompt statement ( $p_{icl}$) that we used for the in-context learning.

\begin{table*}[]
    \centering
    \scalebox{0.9}{
    \begin{tabular}{cl}
    \toprule
    Prompt-ID &  Prompt \\ 
    \midrule
    \midrule
$p_{icl}$ & Look at the samples of a sentence and its intelligible paraphrase: \\
& 1. \; \textit{I don't know if you are familiar with that.}  \; =>  \\
& \; \textit{I have no idea if you're familiar with that.} \\
& 2. \; \textit{what other long-range goals do you have besides college?}  \; =>   \\
&  \;  \textit{Apart from college, what are your other long-term objectives?} \\
& 3. \; \textit{I don't have access either. Although, I did at one time}  \; =>   \\
&  \;  \textit{In the past, I had access,  but currently,  I don't.} \\
& 4. \; \textit{Right now I've got it narrowed down to the top four teams.} \; =>   \\
&  \;  \textit{At this point, I've trimmed my options and picked 4 top teams.} \\
& 5. \; \textit{prohibition didn't stop it and didn't do anything really.}  \; =>   \\
&  \;  \textit{It continued despite the prohibition, which didn't accomplish anything.} \\
    \\
& Similarly, generate an \textbf{intelligible} paraphrase for the 
input sentence: \texttt{\{input text\}} \\ 
    \bottomrule
    \end{tabular}}
    \caption{The prompt used for the in-context learning setup.}
    \label{tab: prompt icl}
\end{table*}

\begin{table*}[]
\centering
\setlength{\tabcolsep}{10pt}
\begin{tabular}{l|lllll}
\toprule
Prompt-ID &  STS$\uparrow$ & LD$\uparrow$ & \textit{PWR-}PhLen$\downarrow$ & \textit{PWR-}PPL$\downarrow$ & \cellcolor{LGreen}\textit{PWR-}STOI $\uparrow$ \\ \midrule
$p_{icl}$ &  0.872 & 0.627 & 1.250* & 0.947  & \cellcolor{LGreen}0.997 \\

\bottomrule
\end{tabular}
\caption{An evaluation of the ICL setup. LLM fails to improve acoustic intelligibility (\textit{PWR}-STOI < $1.0$), though it learns to capture the demonstrated textual attributes like lexical deviation and predictability.
}
\label{tab: evaluation on icl}
\end{table*}

\paragraph{Results and Analysis} 
As shown in \mbox{Table \ref{tab: evaluation on icl}}, the model learned to generate paraphrases, similar to those given as examples. Compared to the zero-shot learning with minimal task description ($p_{zsl-low}$), the model in the ICL setup ($p_{icl}$) generated texts that are semantically more similar and lexically less divergent from the input sentences. More interestingly, the model also learned to optimize the desired textual attributes like length (\textit{PWR}-PhLen) and linguistic predictability (\textit{PWR}-PPL) of generated paraphrases, even in the absence of prompt tokens to explicitly control those features.
Nevertheless, the \textbf{demonstrations are still not helpful in controlling the non-textual attribute}. We observed that the acoustic intelligibility scores of output sentences were \textit{not significantly} different from their input sentences (\textit{PWR}-STOI = $0.997$).
Once again, this shows the inability of the LLM to generate acoustically intelligible paraphrases, even though it captures textual attributes from the given exemplars.


\begin{table*}[]
    \centering
    \begin{tabular}{c|l}
         \toprule
         Sentence\_ID & Sentence  \\
         \midrule
         \midrule
         \textit{$s_1$} & they give more information than opinions \\ 
         \textit{$s_2$} & they seem to give more of just the facts than opinions\\ 
         \textit{$s_3$} & they seem to give more facts than opinions \\ 
         \midrule
         \textit{$s_1$} & you don't hear much about it in the big ones
 \\ 
         \textit{$s_2$} & in the big ones you don't hear about it
 \\ 
         \textit{$s_3$} & you never hear about it really in the big ones
 \\ 
         \midrule
         \textit{$s_1$} & I think we talked for a good eight minutes about the subject \\ 
         \textit{$s_2$} & we talked for about eight minutes \\ 
         \textit{$s_3$} & I think we talked for about eight minutes \\ 
         \midrule
         \textit{$s_1$} & I like having people over for dinner \\ 
         \textit{$s_2$} & I enjoy having people over for dinner \\ 
         \textit{$s_3$} & if I have people over for dinner I like it to be
\\ 
         \midrule
         \textit{$s_1$} & I studied every piece of material I could
 \\ 
         \textit{$s_2$} & I studied every part of the material
\\ 
         \textit{$s_3$} & and studied every bit of material that I could study
\\ 
         \midrule
         \textit{$s_1$} & I wanted to be a teacher at one time
\\ 
         \textit{$s_2$} & at one point I wanted to be a teacher
\\ 
         \textit{$s_3$} & I thought at one time I wanted to be a teacher
\\ 
         \midrule
         \textit{$s_1$} & they never imagined it would be a hit
 \\ 
         \textit{$s_2$} & in fact, they never thought it would be a hit
\\ 
         \textit{$s_3$} & they never expected it to be a hit
\\ 
         \midrule
         \textit{$s_1$} & they want a lot more men to participate
\\ 
         \textit{$s_2$} & they need more men to participate
\\ 
         \textit{$s_3$} & they really looking for a lot more men to participate
\\ 
         \midrule
         \textit{$s_1$} & we gave them about seven minutes
\\ 
         \textit{$s_2$} & we gave them about seven minutes according to my watch
\\ 
         \textit{$s_3$} & they were given seven minutes
\\ 
         \midrule
         \textit{$s_1$} & you don't hear much about it in the big ones
\\ 
         \textit{$s_2$} & in the big ones you don't hear about it
\\ 
         \textit{$s_3$} & you never hear about it really in the big ones
\\ 
         \midrule
         \textit{$s_1$} & at that stage of life you only have so much money left
\\ 
         \textit{$s_2$} & you only have a limited amount of money left
\\ 
         \textit{$s_3$} & you only have so much money left at that point in your life
\\ 
         \midrule
         \textit{$s_1$} & I was angry that they were capable of doing that
\\ 
         \textit{$s_2$} & I was mad that they could do that
\\ 
         \textit{$s_3$} & I was just pissed as hell that they could do that
\\ 
         
         \bottomrule
    \end{tabular}
    \caption{A list of paraphrase triplets ($s_1$, $s_2$, $s_3$) from the PiN dataset. Sentences in each triplet are arranged in such a way that $s_1$ is acoustically less intelligible than $s_2$, and acoustically more intelligible than $s_3$, in a listening condition with babble noise at SNR~$-5$ dB.}
    \label{tab: more_examples_PiN_dataset}
\end{table*}

\end{document}